\documentclass[10pt,twocolumn,letterpaper]{article}

\usepackage{iccv}
\usepackage{times}
\usepackage{epsfig}
\usepackage{graphicx}
\usepackage{amsmath}
\usepackage{amssymb}
\usepackage{booktabs}
\usepackage{multicol}
\usepackage{multirow}
\usepackage{caption}
\usepackage[title]{appendix}%

\usepackage[pagebackref=true,breaklinks=true,letterpaper=true,colorlinks,bookmarks=false]{hyperref}

\iccvfinalcopy

\ificcvfinal\fi

\begin{document}

\title{MoDA: Modeling Deformable 3D Objects from Casual Videos}

\author{
Chaoyue Song\textsuperscript{1}, Jiacheng Wei\textsuperscript{1}, Tianyi Chen\textsuperscript{1,2}, Yiwen Chen\textsuperscript{1}, Chuan-Sheng Foo\textsuperscript{3}, Fayao Liu\textsuperscript{3}, Guosheng Lin\textsuperscript{1}\thanks{Corresponding author}\\
\textsuperscript{1}{Nanyang Technological University, Singapore} \\
\textsuperscript{2}{South China University of Technology, China} \\ 
\textsuperscript{3}{Institute for Inforcomm Research, A*STAR, Singapore} \\ 
{\tt\small \{chaoyue002@e., gslin@\}ntu.edu.sg}
}

\ificcvfinal\thispagestyle{empty}\fi

\twocolumn[{%
\renewcommand\twocolumn[1][]{#1}%
\maketitle
\begin{center}
    \centering
    \captionsetup{type=figure}
    \includegraphics[scale=0.22]{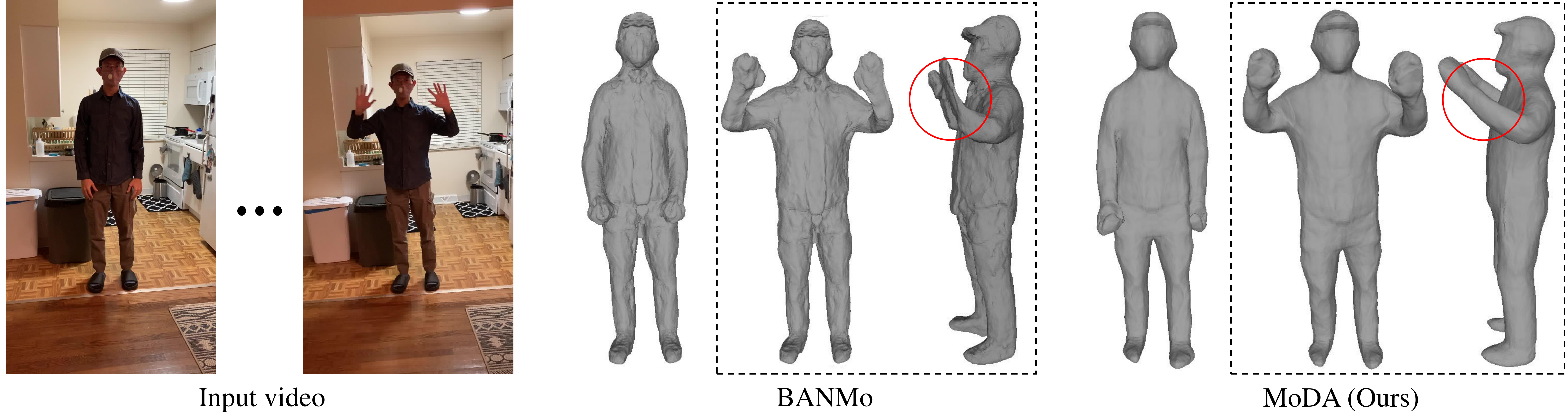}
    \captionof{figure}{In this work, we introduce MoDA that can reconstruct deformable 3D objects from the input casual videos with neural deformation models. Deformation models are used to transform 3D points between the canonical space (rest pose) and the observation space (deformed pose). Previous work BANMo \cite{yang2022banmo} uses linear blend skinning as their deformation model, resulting in visible skin-collapsing artifacts on the arms. MoDA can solve this problem with the proposed neural dual quaternion blend skinning.}
    \label{fig1}
\end{center}%
}]
\begin{abstract}
In this paper, we focus on the challenges of modeling deformable 3D objects from casual videos. With the popularity of neural radiance fields (NeRF), many works extend it to dynamic scenes with a canonical NeRF and a deformation model that achieves 3D point transformation between the observation space and the canonical space. Recent works rely on linear blend skinning (LBS) to achieve the canonical-observation transformation. However, the linearly weighted combination of rigid transformation matrices is not guaranteed to be rigid. As a matter of fact, unexpected scale and shear factors often appear. In practice, using LBS as the deformation model can always lead to skin-collapsing artifacts for bending or twisting motions. To solve this problem, we propose neural dual quaternion blend skinning (NeuDBS) to achieve 3D point deformation, which can perform rigid transformation without skin-collapsing artifacts. In the endeavor to register 2D pixels across different frames, we establish a correspondence between canonical feature embeddings that encodes 3D points within the canonical space, and 2D image features by solving an optimal transport problem. Besides, we introduce a texture filtering approach for texture rendering that effectively minimizes the impact of noisy colors outside target deformable objects. Extensive experiments on real and synthetic datasets show that our approach can reconstruct 3D models for humans and animals with better qualitative and quantitative performance than state-of-the-art methods. Project page: \url{https://chaoyuesong.github.io/MoDA}.
\end{abstract}

\section{Introduction}\label{sec1}

Modeling deformable 3D objects from casual videos has many potential applications in virtual reality, 3D animated movies, and video games. With the popularity of 2D content creation based on advanced techniques, the demand for 3D content creation \cite{Wei_2023_CVPR, chen2024sculpt3d} is becoming more and more urgent for users. Recent works for rigid objects \cite{novotny2017learning, henzler2021unsupervised} cannot generalize to deformable object categories such as humans and animals, which tend to be the focus of content creation today.
Others requiring synchronized multi-view video inputs \cite{peng2021neural, peng2021animatable} are also not available to general users. Therefore, we focus on the challenges of learning deformable 3D objects from casually collected videos in this work. To achieve this goal, we need to learn how to represent deformable objects and model their articulated motions from videos.

Neural radiance field (NeRF) \cite{mildenhall2021nerf} is proposed as a representation of static 3D scenes into volume density and view-dependent radiance. It has shown impressive performance with volume rendering techniques. To extend NeRF to dynamic scenes, recent methods \cite{pumarola2021d, park2021nerfies, park2021hypernerf, yang2022banmo, peng2021animatable} introduce a canonical neural radiance field that models the shape and appearance, and a deformation model that achieves 3D point transformation between the observation space and the canonical space. NSFF\cite{li2021neural} and D-NeRF\cite{pumarola2021d} propose a displacement field to perform the deformation. Nerfies \cite{park2021nerfies} and HyperNeRF \cite{park2021hypernerf} represent their deformation model as a dense $\mathrm{SE(3)}$ field. These methods fail when the
motion between deformable objects and the background is large. Recently, BANMo \cite{yang2022banmo} achieves their deformation model via linear blend skinning (LBS) to solve this problem. However, the linearly weighted combination of rigid transformation matrices is not necessarily a rigid transformation. As a matter of fact, unexpected scale and shear factors always appear. In practice, using LBS as the deformation model can always lead to skin-collapsing artifacts for bending or twisting motions as shown in Figure \ref{fig1}.

In this work, we present \textbf{MoDA} to \textbf{Mo}del \textbf{D}eform\textbf{A}ble 3D objects from multiple casual videos. To handle the large motions between deformable objects and the background without introducing skin-collapsing artifacts, we propose neural dual quaternion blend skinning (NeuDBS) as our deformation model to achieve the observation-to-canonical and canonical-to-observation transformation, which can guarantee the transformations are rigid by blending unit dual quaternions \cite{kavan2007skinning}. With a canonical NeRF as our shape and appearance model, we achieve rigid articulated motions with the proposed deformation model. In the endeavor to register 2D pixels across different frames, we establish a correspondence between canonical feature embeddings that encodes 3D points within the canonical space, and 2D image features. To further promote a one-to-one matching process, we have structured the learning of 2D-3D correspondence learning as an optimal transport problem. Besides, we introduce a texture filtering approach for texture rendering that effectively minimizes the impact of noisy colors (e.g., background colors) outside target deformable objects. Extensive experiments on real and synthetic datasets show that MoDA reconstructs 3D deformable objects like humans and animals with better qualitative and quantitative performance than state-of-the-art methods. 

We summarize our main contributions as follows:

$\bullet$ We introduce MoDA to model deformable 3D objects from multiple casual videos. Through extensive experiments, we demonstrate that MoDA has a better performance than state-of-the-art methods quantitatively and qualitatively on several different datasets.

$\bullet$ To handle the large motions between deformable objects and the background without introducing skin-collapsing artifacts, we propose neural dual quaternion blend skinning (NeuDBS) as our deformation model to transform the 3D points between observation space and canonical space.

$\bullet$ To register 2D pixels across different frames, we establish the correspondence between canonical feature embeddings of 3D points in the canonical space and 2D image features by solving an optimal transport problem.

$\bullet$ We design a texture filtering approach for texture rendering that effectively minimizes the impact of noisy colors outside target deformable objects.

\section{Related work}
\subsection{3D human and animal models}
Many methods rely on parametric shape models \cite{loper2015smpl, pavlakos2019expressive, vo2020spatiotemporal, xiang2019monocular, zuffi20173d, zuffi2018lions, mahmood2019amass} to reconstruct 3D human and animal. These parametric models are constructed from registered 3D scans of humans or animals. They are popular in building 3D shapes from images or videos \cite{badger20203d, biggs2020left, kocabas2020vibe, zuffi2019three} and 3D human or animal generation tasks \cite{song20213d, 10076900, yang2023attrihuman}. Recently, the human model (SMPL) is also used to learn skinning weights for linear blend skinning in \cite{peng2021animatable}. Nonetheless, constructing parametric models for certain categories, such as various types of animals, proves to be challenging due to the difficulty in obtaining a sufficient amount of data.

\subsection{3D reconstruction from images or videos}
There are many prior methods \cite{kanazawa2018learning, goel2020shape, li2020online, li2020self, wu2021dove, ye2021shelf} learn 3D reconstruction from images or videos with the supervision of 2D annotations (key points, optical flow, etc). Their performance will usually be limited due to their reliance on rough shape templates. Neural implicit surface representations \cite{ oechsle2021unisurf, saito2019pifu, saito2020pifuhd, shi2022garf, shi2023usr, wang2021neus, yariv2020multiview} also have many applications in image or video reconstruction. \cite{novotny2017learning, henzler2021unsupervised} learn to reconstruct rigid objects from videos. In this work, we focus on the deformable categories, e.g., humans and animals. Recent work, such as LASR \cite{yang2021lasr} and ViSER \cite{yang2021viser}, optimizes a single 3D deformable model on a monocular video using the mask and optical flow supervision. However, they always introduce unrealistic articulated motions. With the popularity of neural radiance fields \cite{mildenhall2021nerf}, there are many works \cite{liu2021neural, peng2021neural, peng2021animatable, noguchi2021neural, su2021nerf, weng2022humannerf, neural-human-radiance-field,park2021hypernerf,park2021nerfies,yu2021pixelnerf, chen2021mvsnerf, song2022nerfplayer, listreaming, li2022tava} learning to reconstruct the shape and appearance from images or videos with a NeRF-based template. Instead of learning the density in NeRF directly, we use the Signed Distance Function (SDF) that has a well-defined surface at the zero level-set.

\subsection{Neural radiance fields for dynamic scenes}
Recently, many works represent dynamic scenes by learning a deformation model to map the observed points to a canonical space. NSFF\cite{li2021neural} introduces scaled scene flow to displace the 3D points, D-NeRF \cite{pumarola2021d} learns a displacement to transform the given point to the canonical space, NR-NeRF \cite{tretschk2021non} learns a rigidity network to model the deformation of non-rigid objects. Nerfies and HyperNeRF \cite{park2021nerfies, park2021hypernerf} learn a dense $\mathrm{SE(3)}$ field to formulate the deformation. These methods always fail when the motion between deformable objects and the background is large. \cite{liu2021neural, peng2021neural, peng2021animatable, noguchi2021neural, su2021nerf, weng2022humannerf, neural-human-radiance-field} were proposed to solve this problem. However, they either rely on a 3D human model (e.g., SMPL \cite{loper2015smpl}) or synchronized multi-view videos. BANMo \cite{yang2022banmo} can build 3D shapes from casual videos without human or animal models, it learns the deformation model using linear blend skinning (LBS), which always produces skin-collapsing artifacts. To mitigate this issue, we model the deformation with neural dual quaternion blend skinning (NeuDBS) in this work.

\subsection{Correspondence Learning}
Several prior works\cite{yang2021viser, yang2022banmo} have utilized soft-argmax regression to establish a correlation between a canonical feature embedding, which encodes semantic information of three-dimensional points in the canonical space, and two-dimensional pixel features. However, soft-argmax matching, which computes cosine similarities \cite{Yang_2021_CVPR}, can result in many-to-one matching problems. Optimal transport has emerged as an influential tool in addressing this issue, particularly due to its propensity to promote one-to-one matching. This approach has been used in scene flow prediction between point clouds \cite{puy2020flot, Li_2021_CVPR}, 3D semantic segmentation \cite{Shi_2022_CVPR} and few-shot segmentation \cite{liu2022few}. Other research, such as those conducted by Song et al. \cite{song20213d, 10076900}, has applied optimal transport to learn the correspondence between different meshes for 3D pose transfer. In this work, we extend this line of investigation by solving an optimal transport problem to build the correspondence between canonical feature embeddings and 2D pixel features, enabling the registration of pixel observations across varying frames.

\begin{figure}
  \centering
  \includegraphics[scale=0.187]{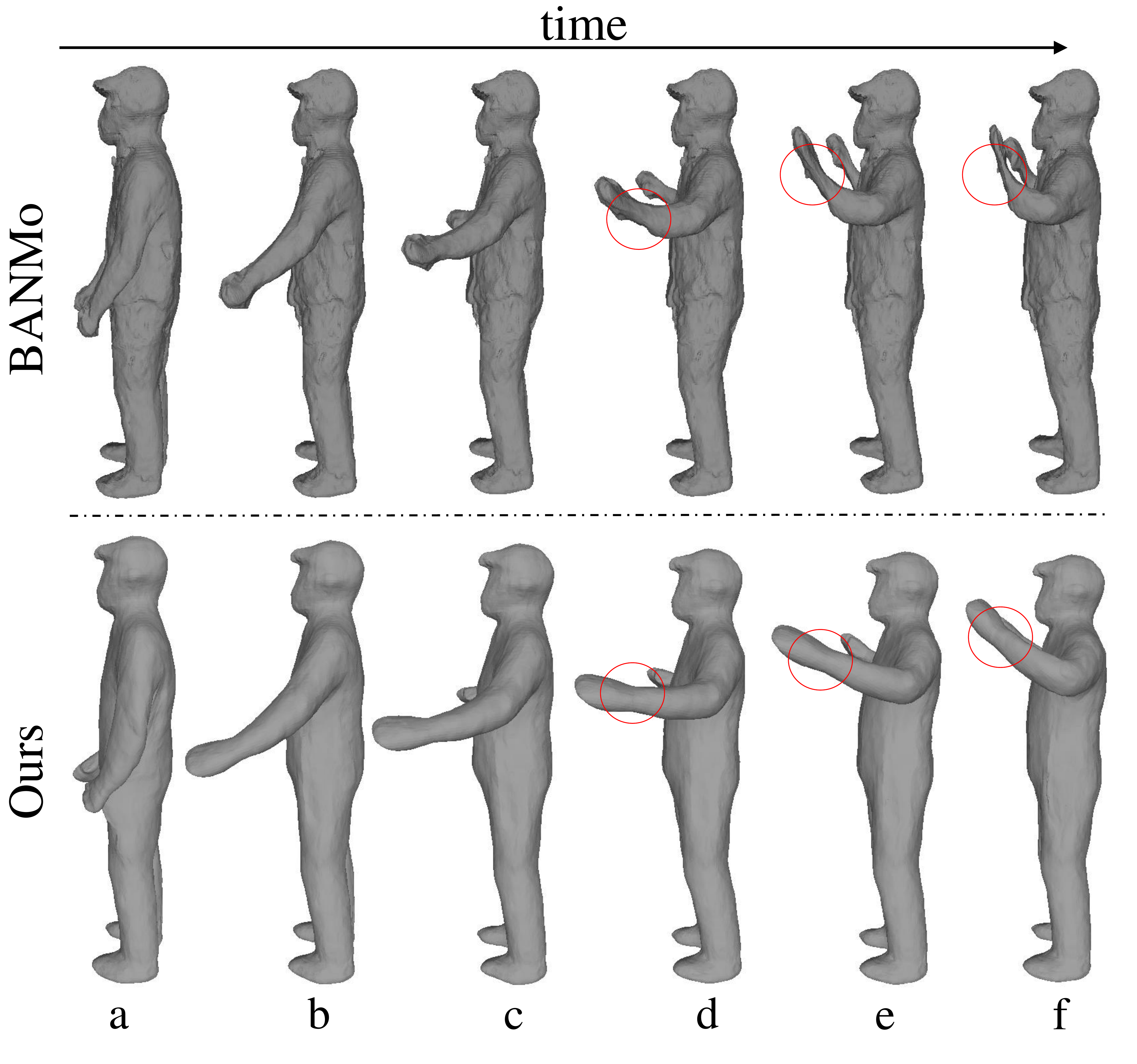}
  \caption{From state \textit{a} to \textit{c}, BANMo and our method can both perform well for motion with small joint rotations. From state \textit{d} to \textit{f}, BANMo has more and more obvious skin-collapsing artifacts for motion with large rotations, our method resolves the artifacts with the proposed NeuDBS.}
  \label{fig2}
\end{figure}
\section{Revisit linear blend skinning}
\label{bk}
The aim of linear blend skinning (LBS) \cite{lewis2000pose, jacobson2014skinning} is to blend transformation matrices linearly and then transform vertices in the rest pose to the expected position in the deformed pose. Each vertex in the mesh can be influenced by multiple joints. The influence of joints on each vertex is controlled by skinning weights. We assume that vertex $\mathbf{v}$ is influenced by joints $\{j_{1},..., j_{n}\}$ with skinning weights $\{w_{1}, ... ,w_{n}\}$. 
Then the transformed vertex position can be formulated as
\begin{equation}
    \mathbf{v}^{\prime} = (\sum_{i=1}^{n}{w_{i}T_{i}})\mathbf{v},
\end{equation}
where $T_{i} \in \mathrm{SE}(3)$ is the transformation matrix. Although each $T_{i}$ represents a rigid transformation, the linearly weighted combination of them is not necessarily a rigid transformation since the addition of orthonormal matrices is not closed. Scale and shear factors always appear. Therefore, the blended transformation matrix applied to vertices tends to cause the limb to shrink and lose volume for bending and twisting motions, which is known as skin-collapsing artifacts. We refer readers to \cite{kavan2007skinning} for details.

LBS has shown impressive performance as the deformation model to represent dynamic scenes in BANMo \cite{ yang2022banmo},
but it still has obvious limitations as discussed above.  
To better understand the performance of LBS as the deformation model, we compare BANMo and our method on a relatively complete human motion sequence from state \textit{a} to \textit{f} in Figure \ref{fig2}. For BANMo (the first line in Figure \ref{fig2}), it performs well and has no obvious skin-collapsing artifacts for motion with small joint rotations (state \textit{a} to \textit{c}), but the artifacts are more and more clear for larger rotations (state \textit{d} to \textit{f}). Our method (the second line in Figure \ref{fig2}) can solve this problem using neural dual quaternion blend skinning (NeuDBS) which will be described in Section \ref{deformodel}.

\begin{figure*}
  \centering
  \includegraphics[scale=0.355]{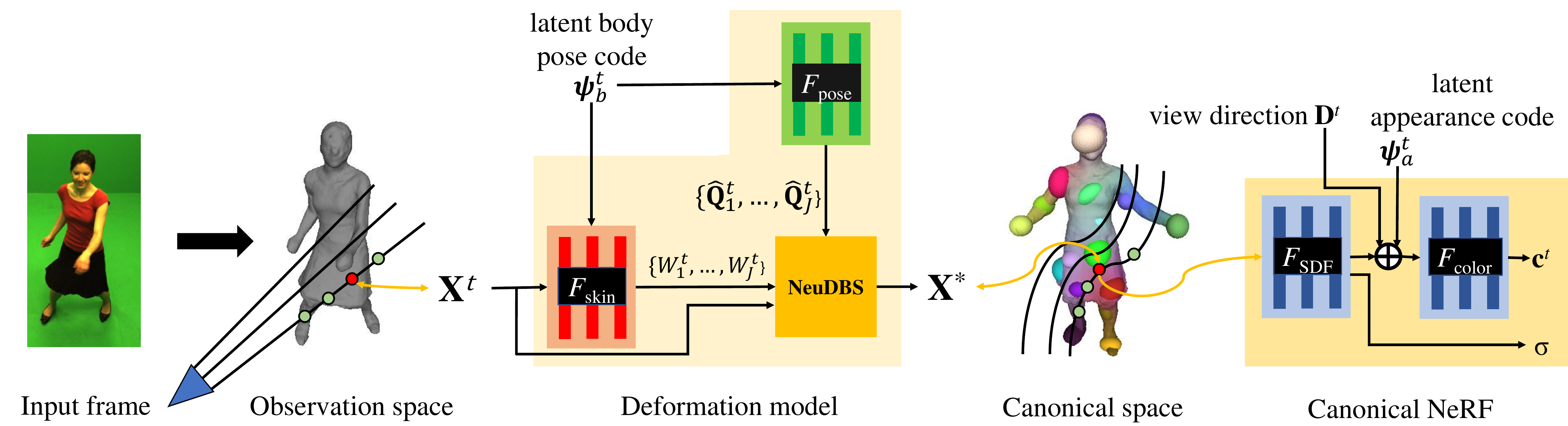}
  \caption{\textbf{The overview of MoDA.} We represent the deformable 3D objects from multiple casual videos with a shape and appearance model based on a canonical neural radiance field and a deformation model that achieves 3D point transformation between the observation space and the canonical space. Instead of linear blend skinning used in previous works, we propose NeuDBS as our deformation model. With the learned unit dual quaternions and the skinning weights, we can transform $\mathbf{X}^{t}$ from the observation space to $\mathbf{X}^{*}$ in the canonical space. We visualize the joints and the skinning weights (as surface colors) in the canonical space.}
  \label{fig3}
\end{figure*}

\section{Method}

The overview of our approach is shown in Figure \ref{fig3}. In this work, we represent the deformable 3D objects from multiple casual videos with a shape and appearance model (Section \ref{sec41}) based on a canonical neural radiance field and a deformation model (Section \ref{deformodel}) that transforms 3D points between the observation space and the canonical space. To resolve the skin-collapsing artifacts seen in previous methods, we propose neural dual quaternion blend skinning (NeuDBS) to achieve the expected rigid transformations by blending unit dual quaternions. In order to register 2D pixels across different frames, we formulate the correspondence learning between canonical feature embeddings of 3D points in the canonical space and 2D image features as an optimal transport problem (Section \ref{sec43}). Furthermore, we design a texture filtering approach (Section \ref{sec44}) for texture rendering that effectively minimizes the impact of noisy colors outside target deformable objects.
\subsection{Shape and appearance model}
\label{sec41}
We first introduce how to model the shape and appearance of a deformable object in canonical space. 
As in Neural Radiance Fields (NeRF) \cite{mildenhall2021nerf}, we learn the color and density of a 3D point $\mathbf{X}^{*} \in \mathbb{R}^{3}$ in the canonical space,

\begin{equation}
    \mathbf{c}^{t} = F_{\mathrm{color}}(\mathbf{X}^{*}, \mathbf{D}^{t}, \boldsymbol{\psi}_{a}^{t}),
    \label{color}
\end{equation}
\begin{equation}
    \sigma = \Phi_{\beta}(F_{\mathrm{SDF}}(\mathbf{X}^{*})),
    \label{density}
\end{equation}
where $F_{\mathrm{color}}$ and $F_{\mathrm{SDF}}$ are MLP networks, $\mathbf{D}^{t} = (\phi^{t}, \theta^{t})$ is the time-varying view direction and $\boldsymbol{\psi}_{a}^{t}$ is a 64-dimensional latent appearance code to encode the appearance variations.

Our canonical shape is modeled by $F_{\mathrm{SDF}}$, which predicts signed distances for 3D points in the canonical space. To perform volume rendering as \cite{mildenhall2021nerf}, we need to convert signed distances into density. In this work, we use the Cumulative Distribution Function of the Laplace distribution with zero mean and $\beta$ scale, denoted as $\Phi_{\beta}(\cdot)$. $\beta$ is a learnable parameter. As discussed in \cite{wang2021neus, yariv2021volume}, the Signed Distance Function (SDF) has a well-defined surface at the zero level-set compared with the density function used in NeRF.

\subsection{Deformation model}
\label{deformodel}
In this section, we introduce how to achieve the 3D point transformation between the observation space and the canonical space via the deformation model. We denote the canonical-to-observation and observation-to-canonical deformation as $\mathcal{D}^{t}_{c\xrightarrow{}o}$ and $\mathcal{D}^{t}_{o\xrightarrow{}c}$ respectively. The body pose in the canonical space is also known as \textit{rest pose}.

\noindent\textbf{Canonical-Observation deformation.}
For any 3D point $\mathbf{X}^{t}$ in the observation space, we can map it to the corresponding point $\mathbf{X}^{*}$ in the canonical space via the deformation model. Here, we denote $\mathbf{C}^{t} \in \mathrm{SE}(3)$ as the transformation of the camera pose from the canonical space to the observation space at time $t$, and $\hat{\mathbf{Q}}^{t}_{j} \in \mathbb{R}^{8}$ as the rigid transformation represented by a unit dual quaternion that transforms the $j$-th joint from the rest pose to the deformed position at time $t$, then
\begin{equation}
    \mathbf{X}^{t} = \mathcal{D}^{t}_{c\xrightarrow{}o}(\mathbf{X}^{*}) = \mathbf{C}^{t}\hat{\mathbf{Q}}^{t}_{c\xrightarrow{}o}\mathbf{X}^{*},
\end{equation}
\begin{equation}
    \mathbf{X}^{*} = \mathcal{D}^{t}_{o\xrightarrow{}c}(\mathbf{X}^{t}) = \hat{\mathbf{Q}}^{t}_{o\xrightarrow{}c}(\mathbf{C}^{t})^{-1}\mathbf{X}^{t},
\end{equation}
where $\hat{\mathbf{Q}}^{t}_{o\xrightarrow{}c}$ and $\hat{\mathbf{Q}}^{t}_{c\xrightarrow{}o}$ are blended by 
$J$ rigid transformations that transform the joints between the rest pose and the deformed positions at time $t$. Multiplication between dual quaternions and 3D coordinates can be done by simply converting the 3D coordinates to dual quaternion format. 

Transformations for body pose and camera pose are respectively parametrized by MLP networks $F_{\mathrm{pose}}$ and $F_{\mathrm{cam}}$,
\begin{equation}
    \hat{\mathbf{Q}}^{t} = F_{\mathrm{pose}}(\boldsymbol{\psi}_{b}^{t}), \quad \mathbf{C}^{t} = F_{\mathrm{cam}}(\boldsymbol{\psi}_{c}^{t})\mathbf{C}_{0}^{t},
\end{equation}
where $\hat{\mathbf{Q}}^{t} = \{\mathbf{\hat{Q}}_{1}^{t},...,\mathbf{\hat{Q}}_{J}^{t}\}$ are the learned unit dual quaternions for rigid body pose transformations. $\mathbf{C}_{0}^{t}$ is the initial camera pose learned from \textit{PoseNet} \cite{yang2022banmo,zhang2021ners}. $\boldsymbol{\psi}_{b}^{t}$ and $\boldsymbol{\psi}_{c}^{t}$ are 128-dimensional latent body pose code and camera pose code respectively.

\noindent\textbf{Neural dual quaternion blend skinning.}Dual quaternion blend skinning (DBS) was first proposed by Kavan \textit{et al.} \cite{kavan2007skinning} and can effectively resolve the skin-collapsing artifacts. All DBS-related parameters (body pose transformation, skinning weights, and joints) are predefined in \cite{kavan2007skinning}, but are unknown and difficult to obtain in our task. Therefore, our main challenge is to define and predict these parameters, the basic idea is that we learn them with MLP networks. 

To prevent the skin-collapsing artifacts as discussed in Section \ref{bk} and better model the deformation of 3D objects, we propose neural dual quaternion blend skinning (NeuDBS) as our deformation model. Instead of blending the transformation matrices, our method blends the unit dual quaternions linearly and then normalizes the results to get the final dual quaternion. Given learned unit dual quaternions $\{\mathbf{\hat{Q}}_{1}^{t},...,\mathbf{\hat{Q}}_{J}^{t}\}$ that represent the rigid transformations, the blended unit dual quaternion can be computed as follows,
\begin{equation}
   \mathbf{\hat{Q}}^{t}_{c\xrightarrow{}o} = \frac{\sum_{j=1}^{J}W_{j, c\xrightarrow{}o}^{t}\mathbf{\hat{Q}}_{j}^{t}}{\left\|\sum_{j=1}^{J}W_{j, c\xrightarrow{}o}^{t}\mathbf{\hat{Q}}_{j}^{t}\right\|}, 
\end{equation}
\begin{equation}
    \mathbf{\hat{Q}}^{t}_{o\xrightarrow{}c} = \frac{\sum_{j=1}^{J}W_{j, o\xrightarrow{}c}^{t}(\mathbf{\hat{Q}}_{j}^{t})^{-1}}{\left\|\sum_{j=1}^{J}W_{j, o\xrightarrow{}c}^{t}(\mathbf{\hat{Q}}_{j}^{t})^{-1}\right\|},
\end{equation}
where $W_{j, c\xrightarrow{}o}^{t}$ and $W_{j, o\xrightarrow{}c}^{t}$ are skinning weights that control the influence of $j$-th joint on $\mathbf{X}^{*}$ and $\mathbf{X}^{t}$ respectively.
By computing a unit dual quaternion, NeuDBS always returns a valid rigid transformation which can prevent the skin-collapsing artifacts.

In addition to the difference in parameter definition, Kavan \textit{et al.} must convert predefined skinning matrices ($3 \times 3$ rotation and $1 \times 3$ translation) to dual quaternions first to apply DBS.
MoDA learns $7$ scalars per joint via MLP and then directly derives dual quaternions (see next paragraph for details), which does not require the quaternion-matrix conversion and is more efficient. We will also introduce how to define and optimize the skinning weights in the supplement.

\noindent\textbf{Learning of body pose transformation.} In BANMo \cite{yang2022banmo}, they learn $6$ scalars per joint for body pose transformation. $3$ of them are for translation, the other $3$ scalars are the logarithmic representation of the rotation matrix. They convert logarithmic representations of rotation matrices into $3 \times 3$ rotation matrices. Therefore, BANMo ends up requiring a $3 \times 4$ matrix per joint for body pose transformation.

In this work, we learn $7$ scalars per joint via $F_{\mathrm{pose}}$ for the rigid transformation. $3$ of them ($t_{1}, t_{2}, t_{3}$) are for translation, then we can obtain the quaternion $\mathbf{q}_{t} = [0, t_{1}, t_{2}, t_{3}]$ representing the translation, where $0$ is the scalar part, $t_{1}\mathbf{i}+t_{2}\mathbf{j}+t_{3}\mathbf{k}$ is the vector part of a quaternion.
The other $4$ scalars are the quaternion $\mathbf{q}_{r}$ that represents the rotation. Then we convert them to the dual quaternion,
\begin{equation}
    \mathbf{Q}_{r} = \hat{\mathbf{q}}_{r}, \quad
    \mathbf{Q}_{d} = \frac{1}{2}\mathbf{q}_{t}\bigotimes\hat{\mathbf{q}}_{r},
\end{equation}
where $\mathbf{Q}_{r}$ is the real part of the dual quaternion, $\mathbf{Q}_{d}$ is the dual part of the dual quaternion. $\hat{\mathbf{q}}_{r}$ is the normalized quaternion to make sure the calculated dual quaternion is the unit dual quaternion.   $\bigotimes$ means the multiplication between quaternions.

In sum, BANMo learns $6$ scalars and requires a $3 \times 4$ matrix per joint to represent the body pose transformation. Our MoDA learns $7$ scalars but only requires $8$ scalars per joint for the body pose transformation, which introduces a more efficient representation. Linear Blend Skinning (LBS) \cite{jacobson2014skinning, lewis2000pose} is widely recognized for its efficiency, and the training time required for MoDA with NeuDBS is comparable to that of BANMo, please refer to the appendix for details.

\subsection{2D-3D matching via optimal transport}
\label{sec43}
To match pixels at different frames, we establish the correspondence between canonical feature embeddings of 3D points in the canonical space and 2D image features. In \cite{yang2021viser, yang2022banmo}, they employ soft-argmax regression to learn the 2D-3D correspondence by calculating the cosine similarity. To perform better matching, we formulate the 2D-3D matching as an optimal transport problem that encourages one-to-one matching. Given pixels $\{\mathbf{x}^{t}\}_{N_{pixel}}$ and 3D points $\{\mathbf{X}^{*}\}_{N_{point}}$ in the canonical space, we learn the pixel features $f_{pixel}(\mathbf{x}^{t}) \in \mathbb{R}^{16 \times N_{pixel}}$ with CSE \cite{neverova2020continuous, neverova2021discovering} and the canonical feature embeddings $f_{point}(\mathbf{X}^{*}) = F_{emb}(\mathbf{X}^{*}) \in \mathbb{R}^{16 \times N_{point}}$. We obtain the correlation matrix $\mathbf{M} \in \mathbb{R}^{N_{pixel} \times N_{point}}$ by calculating their cosine similarity, 
\begin{equation}
    \mathbf{M}(j,k) = \frac{f_{pixel}(j)^{\top}f_{point}(k)}{\left\|f_{pixel}(j)\right\|\left\|f_{point}(k)\right\|} 
\end{equation}
where $\mathbf{M}(j,k)$ is the matching score between $j$-th pixel and $k$-th point. To formulate the optimal transport problem, we define a matching matrix $\mathbf{T} \in \mathbb{R}^{N_{pixel} \times N_{point}}$ and the cost matrix $\mathbf{Z} = 1-\mathbf{M}$. Our goal is to minimize the total cost to get the optimal matching matrix:

\begin{equation}
\begin{aligned}
    \mathbf{T}^{*} = \mathop{\arg\min}_{\mathbf{T}} \sum_{j k} & \mathbf{Z}(j, k)\mathbf{T}(j, k) \\
    s.t. \quad \mathbf{T}\mathbf{1}_{N_{point}} = & \mathbf{1}_{N_{pixel}}N^{-1}_{pixel},  \\ \mathbf{T}^{\top}\mathbf{1}_{N_{pixel}} = & \mathbf{1}_{N_{point}}N^{-1}_{point}.
    \label{tm}
\end{aligned}
\end{equation}

This optimal transport problem can be solved by the Sinkhorn-Knopp algorithm \cite{sinkhorn1967diagonal}. Then we can find the 3D surface point in the canonical space matching to $\mathbf{x}^{t}$ by warping the sampled points $\mathbf{V}^{*}$ in a canonical 3D grid, 
\begin{equation}
    \mathbf{\widetilde{X}}^{*}(\mathbf{x}^{t}) = \sum_{\mathbf{X} \in \mathbf{V}^{*}} \mathbf{T}^{*}\mathbf{X},
\end{equation}

Based on the 3D surface points, we have the 2D-3D matching losses.
We first define a point matching loss \cite{yang2022banmo} as
\begin{equation}
    \mathcal{L}_{match} = \sum_{\mathbf{x}^{t}}{\left\|\mathbf{\widetilde{X}}^{*}(\mathbf{x}^{t}) - \mathbf{X}^{*}(\mathbf{x}^{t})\right\|^{2}_{2}},
\end{equation}
where $\mathbf{X}^{*}(\mathbf{x}^{t})$ is calculated from Eq. \ref{eq17}. We also define a projection loss \cite{kulkarni2020articulation, yang2021viser, yang2022banmo} that encourages the image projection after canonical-to-observation deformation of $\mathbf{\widetilde{X}}^{*}(\mathbf{x}^{t})$ to be close to its original 2D coordinates.
\begin{equation}
    \mathcal{L}_{proj} = \sum_{\mathbf{x}^{t}}{\left\|\mathbf{P}^{t}(\mathcal{D}^{t}_{c\xrightarrow{}o}(\mathbf{\widetilde{X}}^{*}(\mathbf{x}^{t})) - \mathbf{x}^{t}\right\|^{2}_{2}},
\end{equation}
where $\mathbf{P}^{t}$ is the projection matrix of a pinhole camera.

\subsection{Volume rendering and optimization}
\label{sec44}
\noindent\textbf{Texture filtering for volume rendering.}
When predicting the object color as Eq. \ref{color}, the training process will inevitably introduce some noisy textures (e.g., background texture in the first row of Figure \ref{fig7}) that do not belong to the target deformable objects. 

Inspired by \cite{wu2022object,zhi2021place}, we develop a texture filtering approach to mitigate this issue. Rather than relying on semantic fields to calculate opacity in \cite{wu2022object,zhi2021place}, we utilize a texture filtering function $s$ for texture rendering, which can exclude the noisy colors outside objects (i.e., remove the estimated color $\mathbf{c}^{t}$ when SDF $d > 0$).

Here, we define $\mathbf{x}^{t} \in \mathbb{R}^{2}$ as the pixel location at time $t$, and $\mathbf{X}_{k}^{t}$ as the $k$-th sampled point along the ray that originates from $\mathbf{x}_{t}$. Then color $\mathbf{c}$ and opacity $\mathbf{o} \in [0, 1]$ are given by:
\begin{equation}
    \mathbf{c}(\mathbf{x}^{t}) =\sum_{k=1}^{N}\tau_{k}(s_{k}\mathbf{c}_{k}^{t}),
\quad
    \mathbf{o}(\mathbf{x}^{t}) = \sum_{k=1}^{N}\tau_{k},
\end{equation}

where $N$ is the number of sampled points, $\tau_{k}=\alpha_{k}{\prod_{i=1}^{k-1}(1-\alpha_{i})}$, $\alpha_{k} = 1 - \mathrm{exp}(-\sigma_{k}\delta_{k})$, $\delta_{k}$ is the distance between the $k$-th sample and the next, and $\sigma_{k}$ is the density in Eq. \ref{density}. Texture filtering function $s$ is defined as
\begin{equation}
    s = \frac{\gamma}{1+e^{\lambda d}},
\end{equation}
which is a scaled sigmoid function based on SDF $d = F_{\mathrm{SDF}}(\mathbf{X}^{*})$. $s$ gives $0$ weights to $\mathbf{c}^{t}$ of the sampled points that are far away from the object (have large positive SDF values) to exclude them in the rendering process. $\gamma$ and $\lambda$ are scale and temperature parameters.

We can also calculate the surface point,
\begin{equation}
    \mathbf{X}^{*}(\mathbf{x}^{t}) = \sum_{k=1}^{N}{\tau_{k}\mathbf{X}_{k}^{*}},
    \label{eq17}
\end{equation}
where $\mathbf{X}_{k}^{*}$ is obtained by applying the deformation $\mathcal{D}^{t}_{o\xrightarrow{}c}$ to the $k$-th 3D point $\mathbf{X}^{t}_{k}$.

\noindent\textbf{Optimization.}
In addition to the 2D-3D matching losses, we incorporate the following reconstruction losses into our model optimization, which are commonly employed in existing methods such as \cite{mildenhall2021nerf, yariv2020multiview, yang2022banmo}:
\begin{equation}
    \mathcal{L}_{rgb} = \sum_{\mathbf{x}^{t}}{\left\|\mathbf{c}(\mathbf{x}^{t}) - \widetilde{\mathbf{c}}(\mathbf{x}^{t})\right\|^{2}}, 
\end{equation}
\begin{equation}
    \mathcal{L}_{sil} = \sum_{\mathbf{x}^{t}}{\left\|\mathbf{o}(\mathbf{x}^{t}) - \widetilde{\mathbf{s}}(\mathbf{x}^{t})\right\|^{2}}, 
\end{equation}

where $\mathcal{L}_{rgb}$ and $\mathcal{L}_{sil}$ are the pixel color loss and silhouette loss respectively.   $\widetilde{\mathbf{c}}$ and $\widetilde{\mathbf{s}}$ are observed pixel color and silhouette. Here, $\widetilde{\mathbf{s}}$ is extracted from off-the-shelf method \cite{kirillov2020pointrend}. Additional losses will be introduced in the supplement, such as flow loss (the estimated flow is obtained from \cite{yang2019volumetric}).

\section{Experiments}

\subsection{Dataset, metrics, and implementation details}

\noindent\textbf{Casual videos.}
To demonstrate the effectiveness of MoDA, we test it on casual videos of humans and animals. The \textit{casual-cat} dataset includes 11 videos (900 frames in total) of a British shorthair cat, which are collected by \cite{yang2022banmo}. The \textit{casual-human} dataset \cite{yang2022banmo} includes 10 videos (584 frames in total). The \textit{casual-adult} dataset includes 10 videos (1000 frames in total). The capture of the videos has no control for camera and object movements. We use the object silhouette and optical flow predicted by \cite{kirillov2020pointrend, yang2019volumetric} respectively.

\noindent\textbf{AMA dataset.}
To evaluate our method quantitatively, we use the Articulated Mesh Animation (AMA) dataset \cite{vlasic2008articulated} that provides ground truth meshes. AMA is collected with the setup consisting of a ring of 8 cameras. We train our models on 2 sets of videos of the same person (\textit{swing} and \textit{samba}, including 2600 frames in total) with ground truth object silhouettes and the optical flow predicted by \cite{yang2019volumetric}. 

\noindent\textbf{Animated objects dataset.} Besides cats and humans, we quantitatively evaluate our method on other deformable categories. We use the animated objects dataset from TurboSquid (known as \textit{eagle} and \textit{hands}). They both include 5 videos with 150 frames per video. We train these two datasets with the ground truth camera poses, ground truth silhouettes and the optical flow predicted by \cite{yang2019volumetric}.

\noindent\textbf{Metrics.}
To compare different methods quantitatively, we use Chamfer distance (CD) \cite{fan2017point} and F-scores as our evaluation metrics. CD is calculated between the point sets of the reconstructed mesh and the ground truth mesh.
($\mathbf{p}, \mathbf{\widetilde{p}}$)
\begin{equation}
    CD(\mathbf{p}, \mathbf{\widetilde{p}}) = \sum_{x \in \mathbf{p}}\min_{y \in \mathbf{\widetilde{p}}}{\left\|x - y\right\|^{2}_{2}} + \sum_{y \in \mathbf{\widetilde{p}}}\min_{x \in \mathbf{p}}{\left\|x - y\right\|^{2}_{2}}.
\end{equation}
For CD, the lower is better. For F-scores, we compare different methods at distance thresholds $d = 2\%$, and the bigger the better when using F-scores.

\noindent\textbf{Implementation details.}
In this work, we set the number of joints to 25. The initialization of them is similar to BANMo \cite{yang2022banmo}, with unit scale, identity orientation, and uniformly distributed centers. The meshes are extracted by running marching cubes on a $256^{3}$ grid. For more implementation details, please refer to our appendix.

\begin{figure*}
  \centering
  \includegraphics[scale=0.16]{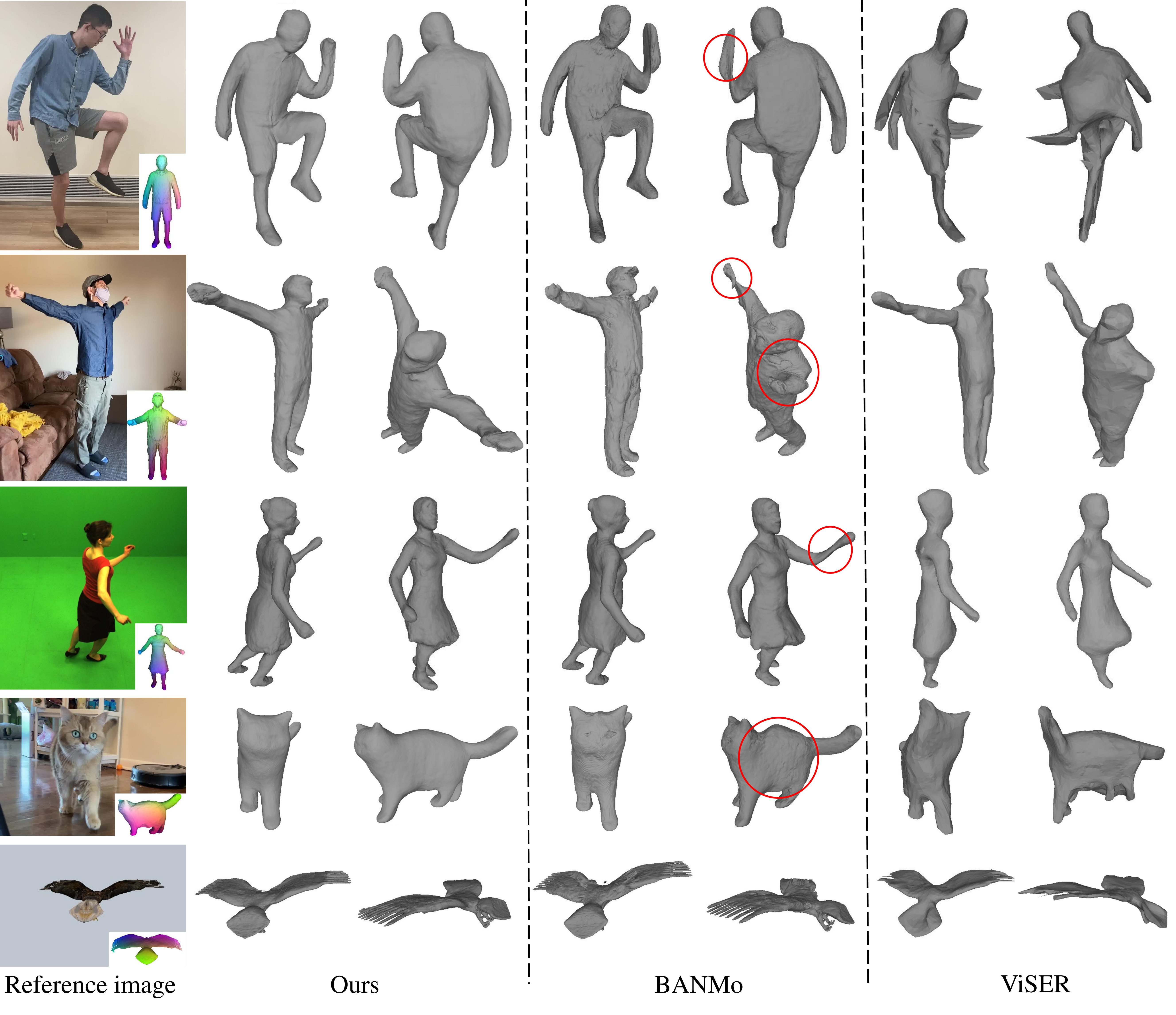}
  \caption{\textbf{Qualitative comparison on multiple videos.} The data is from \textit{casual-adult}, \textit{casual-human}, AMA-\textit{samba}, \textit{casual-cat}, \textit{eagle} from top to bottom. The lower right corner of each reference image is the corresponding rest pose. We show 2 views of the reconstructed results based on the reference images. ViSER \cite{yang2021viser} fails to learn detailed 3D shapes and accurate poses from the videos. BANMo \cite{yang2022banmo} has obvious skin-collapsing artifacts (in the red circles) for motions with large joint rotations while our method performs well. For \textit{eagle} with slight motion, BANMo and our method have close performance.}
  \label{fig4}
\end{figure*}

\begin{table*}
  \centering
  \caption{\textbf{Quantitative comparison between different methods.} We compare our method with state-of-the-art methods on multiple-video and single-video setups. To quantitatively evaluate different methods, we use Chamfer distance (cm, $\downarrow$) and F-score($\%, \uparrow$) as the metrics. Our method has the best performance for both multiple-video and single-video setups.}
  \label{comparison}
  \begin{tabular}{cccccccccc}
    \toprule
  \multirow{2}{*}{Type}  & \multirow{2}{*}{Method} & \multicolumn{2}{c}{AMA-Swing} &  \multicolumn{2}{c}{AMA-Samba} & \multicolumn{2}{c}{Eagle} & \multicolumn{2}{c}{Hands} \\ \cmidrule{3-10}
  & & CD($\downarrow$) & F(2\%, $\uparrow$) &CD($\downarrow$) & F(2\%, $\uparrow$) &CD($\downarrow$) & F(2\%, $\uparrow$) &CD($\downarrow$) & F(2\%, $\uparrow$)  \\
    \midrule
  \multirow{3}{*}{\textit{Multiple}} & ViSER  & 35.8 & 9.9 & 33.8 & 10.0 & 36.9 & 2.5 & 13.5 & 32.6\\
    &  BANMo&  7.9 & 60.8  &  7.9 &  60.4 & 5.2 & 68.7 & 5.1 & 69.5 \\
    & Ours &   \textbf{7.0}  & \textbf{66.3} &  \textbf{5.6}  &  \textbf{75.5} & \textbf{4.8}& \textbf{75.0} & \textbf{4.4} & \textbf{73.9} \\
    \cmidrule{1-10} 
    \multirow{4}{*}{\textit{Single}} &Nerfies  & 39.1   & 5.5 & 42.7 & 4.8 & 27.9 & 10.4 & 33.8 & 6.8 \\
    &HyperNeRF  & 42.9   & 4.9 & 41.7 & 5.2 & 28.6 & 10.2 & 30.9 & 8.3 \\
     &  BANMo &  9.0  & 55.8 & 9.6 & 51.6 &  14.4& 45.1    & 12.7 & 27.2 \\
      &  Ours   &  \textbf{8.5} & \textbf{60.5} & \textbf{8.2}& \textbf{59.7} & \textbf{13.7} & \textbf{45.7}  & \textbf{11.3} & \textbf{32.8}    \\
    \bottomrule
  \end{tabular}
\end{table*}

\begin{figure*}
  \centering
  \setlength{\abovecaptionskip}{0.cm}
  \includegraphics[scale=0.16]{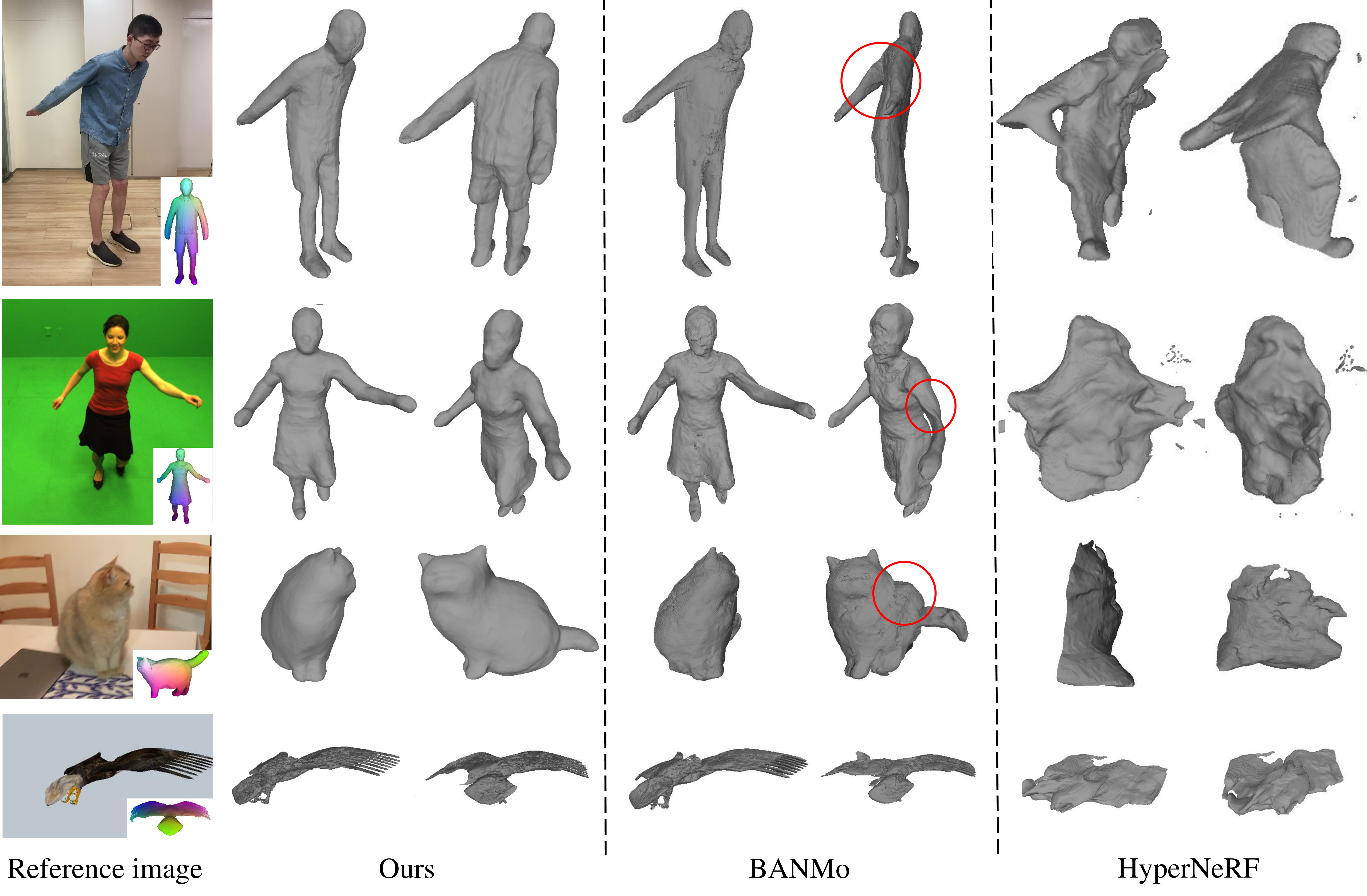}
  \caption{\textbf{Qualitative comparison on a single video.} The data is \textit{casual-adult}, AMA-\textit{swing}, \textit{casual-cat}, \textit{eagle} from top to bottom. The lower right corner of each reference image is the corresponding rest pose. We show 2 views of the reconstructed results based on the reference images.
  HyperNeRF \cite{park2021hypernerf} fails to learn reasonable shapes and deformations. For single-video setups, BANMo \cite{yang2022banmo} still has obvious skin-collapsing artifacts (in the red circles) for motions with large joint rotations while our method performs better.
  } 
  \label{fig5}
\end{figure*}

\subsection{Comparison results on multiple videos}
In this section, we compare MoDA with ViSER \cite{yang2021viser} and BANMo \cite{yang2022banmo} over multiple videos. For a fair comparison, we provide them with the same initial camera poses. And we train BANMo and ViSER using the implementations provided by the authors.  

As shown in Figure \ref{fig4}, ViSER cannot learn detailed shapes and accurate poses from the given videos.
To better demonstrate the influence of the deformation model, we also show the corresponding rest pose for each reference image. For motion with large joint rotations, the results of BANMo have obvious skin-collapsing artifacts (as shown in the red circles). The reconstructed shapes tend to shrink and lose volume, e.g., the arms of humans in \textit{casual-adult}, \textit{casual-human} and AMA-\textit{samba}, and the body of the cat in \textit{casual-cat}. Our method can solve these problems and achieve rigid articulated motions. For \textit{eagle}, BANMo and our method have close performance since the motion of the eagle from the rest pose to the deformed pose is relatively slight. We will show more results on multiple-video setups (including the video demonstrations) in the supplement.

For the quantitative results as shown in Table \ref{comparison}, our method also has a better performance than BANMo and ViSER on multiple-video setups.

\subsection{Comparison results on single-video setups}

In this section, we compare MoDA with Nerfies \cite{park2021nerfies}, HyperNeRF \cite{park2021hypernerf} and BANMo \cite{yang2022banmo} over single-video setups. To make a fair comparison, we also provide them with the same initial camera poses. We reproduce Nerfies, HyperNeRF, and BANMo using the implementations provided by the authors. Besides, we provide Nerfies and HyperNeRF with the ground truth object silhouettes of AMA and Animated object datasets to calculate the silhouette losses which can help to improve the performance.

Nerfies and HyperNerf have very close performance so we only show the results of HyperNeRF in Figure \ref{fig5}. HyperNeRF fails to learn reasonable shapes and deformations for deformable 3D objects when the motion between the object and the background is large. The reconstructed results of BANMo still have clear skin-collapsing artifacts as shown in the red circles while MoDA has a better performance and resolves the skin-collapsing artifacts. Obviously, the performance of BANMo and MoDA both degrade with some unexpected artifacts compared to the multiple-video setups. 

For the quantitative results as shown in Table \ref{comparison}, our method has a better performance than Nerfies, HyperNeRF, and BANMo on single-video setups. The quantitative performance of BANMo and our method on single-video setups also degrades compared to the multiple-video setups.

\begin{figure*}
  \centering
  \includegraphics[scale=0.3]{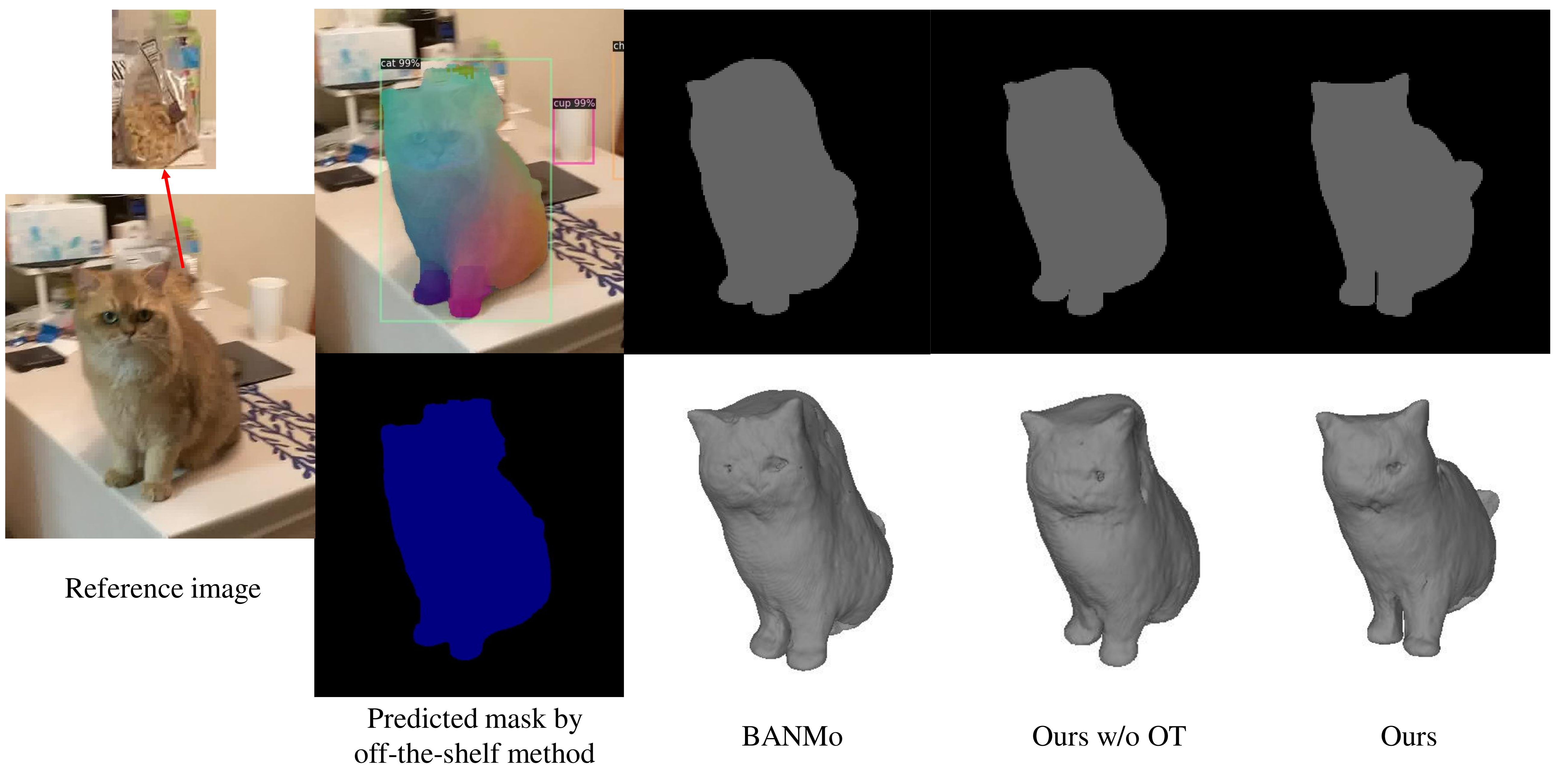}
  \caption{\textbf{Ablation study of optimal transport.} By registering 2D pixels across different frames with optimal transport, we can refine the bad segmentation and predict the consistent 3D shape of the cat.}
  \label{figot}
\end{figure*}

\begin{figure*}
  \centering
  \includegraphics[scale=0.24]{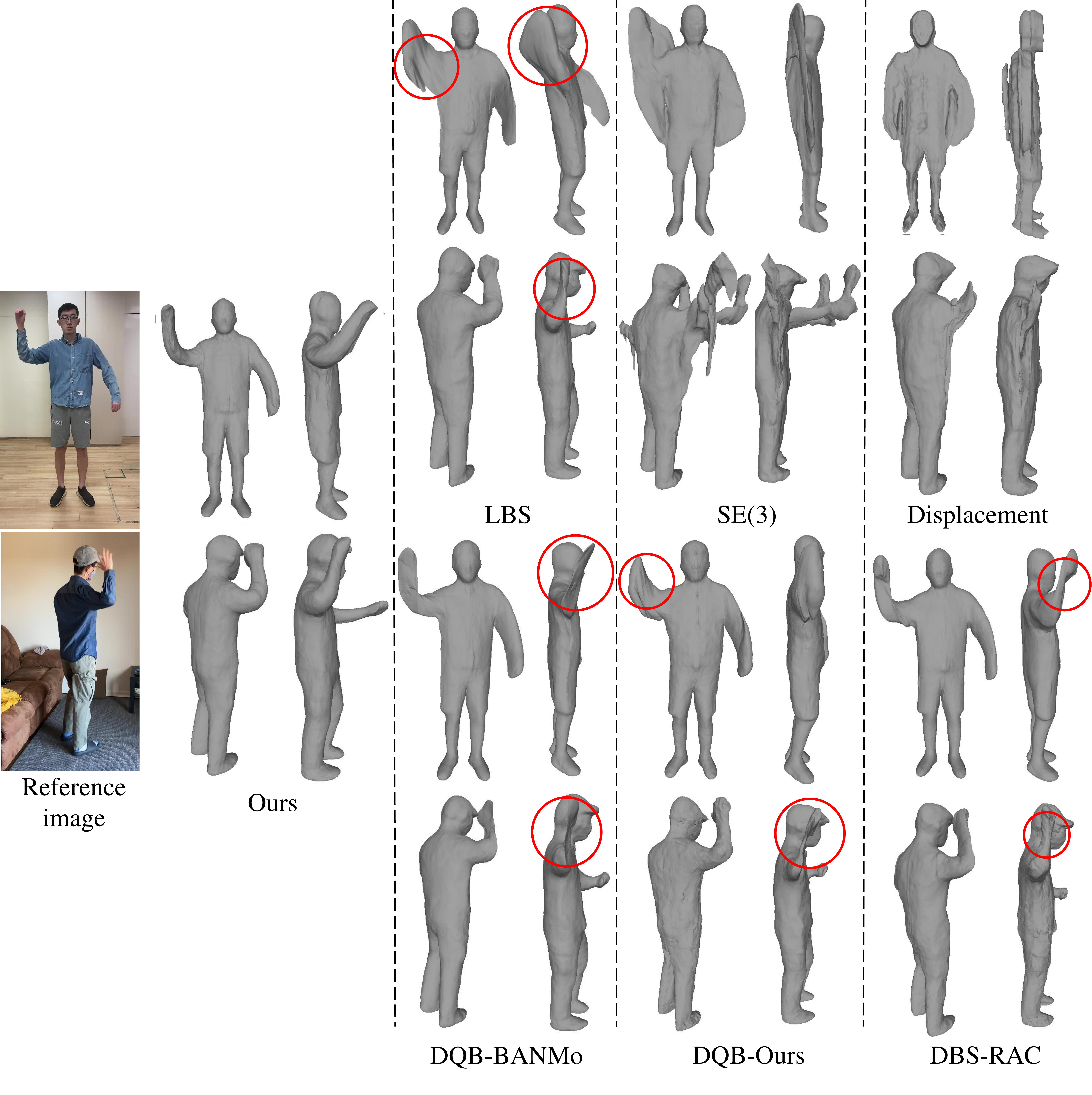}
  \caption{\textbf{Ablation study of deformation models.}  The displacement field and $\mathrm{SE}(3)$ field cannot accurately preserve shapes and learn the pose of humans when dealing with large motions between humans and the background. LBS-based has obvious skin-collapsing artifacts (in the red circles) on the arm during bending motions. Neither DQB-BANMo nor DQB-Ours alleviates these artifacts. Although RAC-DBS demonstrates some improvements over LBS and DQB, artifacts on the arms are still present. Our method achieves the best performance. }
  \label{fig6}
\end{figure*}

\subsection{Ablation study}
\label{ablation}
\noindent\textbf{Optimal transport.} We also evaluate the importance of optimal transport for 2D-3D matching. To disable optimal transport, we use the soft-argmax regression by calculating the cosine similarity that is similar to \cite{yang2021viser, yang2022banmo}. 
According to the results presented in Table \ref{ablationtable}, our method with optimal transport achieves better performance than using soft-argmax regression when testing on AMA dataset \cite{vlasic2008articulated}. In the case of the AMA dataset, which has a clean background and accurate ground truth masks, the addition of optimal transport does not significantly improve performance.

When evaluating our method on the \textit{casual-cat} dataset, the inclusion of optimal transport provides more obvious advantages. It facilitates improved 2D-3D matching, which, in turn, aids in refining the object silhouette. Figure \ref{figot} illustrates this improvement. The predicted mask obtained from \cite{kirillov2020pointrend} is inaccurate due to the close pixel colors between the cat and the snack. In such a scenario, both BANMo \cite{yang2022banmo} and our method without optimal transport struggle to refine the mask and accurately capture the 3D geometry. In most frames, the two objects (cat and snack) are not closely located to the extent of being mistaken for a single object. As a result, by utilizing optimal transport to achieve better matching, we can refine the initial segmentation and predict the consistent 3D shape of the cat.

\begin{figure}[h]
  \centering
  \includegraphics[scale=0.24]{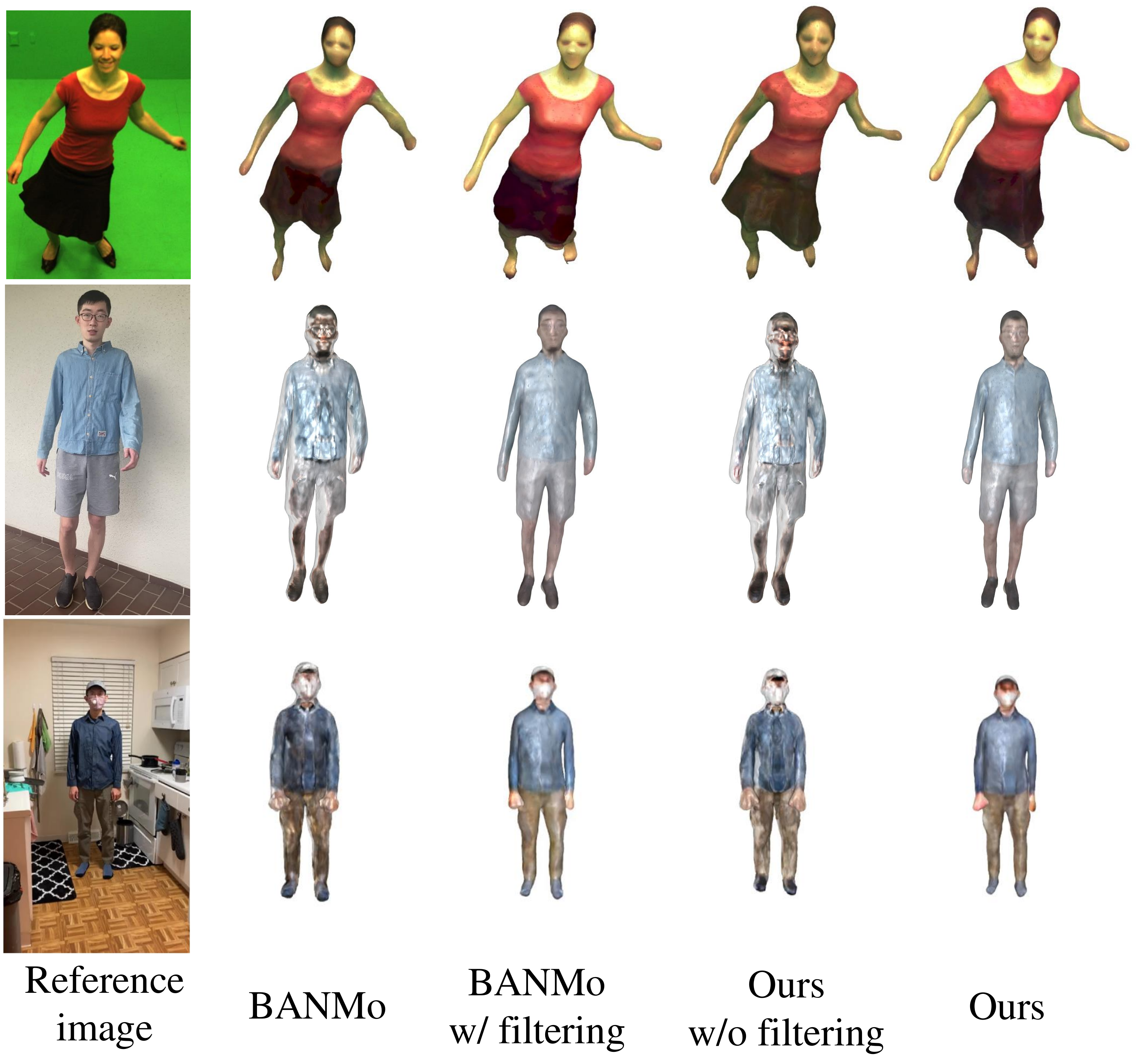}
  \caption{\textbf{Ablation study of texture filtering.}  Texture rendering results of BANMo \cite{yang2022banmo} and MoDA without texture filtering have obvious noisy textures. Adding texture filtering to them can effectively alleviate this issue. }
  \label{fig7}
\end{figure}

\noindent\textbf{Deformation model.} To further validate the effectiveness of our proposed NeuDBS, we compare it with other deformation models. Specifically, we replace NeuDBS with alternative deformation models, including the displacement field from \cite{li2021neural, pumarola2021d}, $\mathrm{SE}(3)$ field from \cite{park2021nerfies, park2021hypernerf} and neural linear blend skinning (LBS) from \cite{yang2022banmo}. We also implement direct quaternion blending (DQB) \cite{ref} which decomposes transformation matrices into quaternions and translations and blends them linearly. To compare with DQB, we design two approaches for learning (translation, quaternion) pairs. \textit{DQB-BANMo}: This method follows BANMo's approach, where we initially learn (translation, logarithmic representation of rotation matrix) pairs and then convert them to (translation, quaternion) pairs. \textit{DQB-Ours}: In alignment with our NeuDBS method, we directly learn (translation, quaternion) pairs. Additionally, we also conduct an ablation study on DBS in RAC \cite{yang2023reconstructing}. They first learn (translation, logarithmic representation of rotation matrix) pairs, following BANMo's approach, and subsequently converting them to dual quaternions.

The qualitative results on \textit{casual-adult} and \textit{casual-human} are shown in Figure \ref{fig6}, the displacement field and $\mathrm{SE}(3)$ field cannot accurately preserve shapes and learn the pose of humans, particularly in the case of large motions. Furthermore, LBS exhibits noticeable skin-collapsing artifacts on the arm during bending motions. Neither DQB-BANMo nor DQB-Ours alleviates these artifacts. Although RAC-DBS demonstrates some improvements over LBS and DQB, artifacts on the arms are still present. In contrast, our proposed NeuDBS demonstrates superior performance compared to these six deformation models. It successfully preserves shapes and avoids skin-collapsing artifacts, resulting in visually improved outputs. To provide quantitative evidence, we present the results on the AMA dataset \cite{vlasic2008articulated} in Table \ref{ablationtable}. Our NeuDBS consistently outperforms the alternative deformation models. Note that while the AMA dataset contains multi-view data, offering richer information, the quantitative distinctions between various deformation models may not be readily apparent. The reconstruction results from casual videos can offer a more straightforward validation of the benefits provided by our NeuDBS.

\begin{table*}
  \centering
  \caption{\textbf{Quantitative ablation studies.} We evaluate different deformation models, the optimal transport module, and texture filtering on AMA dataset. We use Chamfer distance (cm, $\downarrow$) and F-score($\%, \uparrow$) as the metrics.}
  \label{ablationtable}
  \begin{tabular}{cccccc}
    \toprule
  \multirow{2}{*}{Type}  & \multirow{2}{*}{Method} & \multicolumn{2}{c}{AMA-Swing} &  \multicolumn{2}{c}{AMA-Samba}  \\ \cmidrule{3-6}
  & & CD($\downarrow$) & F(2\%, $\uparrow$) &CD($\downarrow$) & F(2\%, $\uparrow$) \\
    \midrule
  \multirow{9}{*}{\textit{Multiple}} & Displacement  & 11.3 & 44.5 & 10.7 & 54.2 \\
  & $\mathrm{SE}(3)$ & 10.4 & 48.2 & 9.4 & 56.4 \\
    &  LBS &  7.6 & 63.7  &  6.7 &  66.9 \\
    &  DQB-BANMo &  7.8 & 62.6  &  6.1 &  69.7\\
    &  DQB-Ours &  7.5 & 63.9  &  6.3 &  68.4\\
    &  DBS-RAC &  7.1 & 65.8  &  6.0 &  73.3\\
    \cmidrule{2-6}
    &  w/o OT &  7.2 & 64.5 &  5.9 &  73.1\\
    & Ours &  \textbf{7.0}  & \textbf{66.3} &  \textbf{5.6}  &  \textbf{75.5}  \\

    \bottomrule
  \end{tabular}
\end{table*}

\begin{figure*}[h]
  \centering
  \includegraphics[scale=0.342]{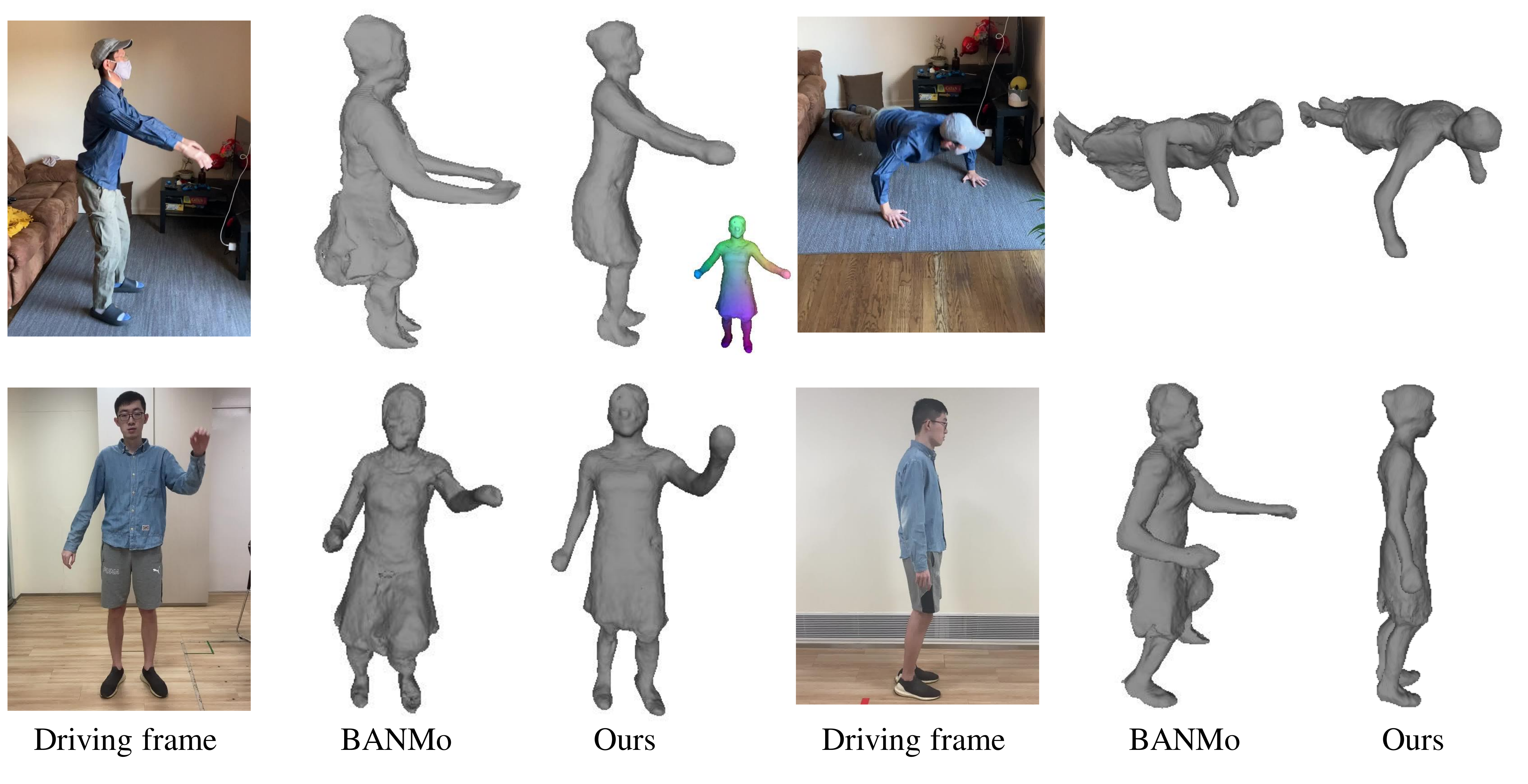}
  \caption{\textbf{Motion re-targeting.}  We compare the motion re-targeting results from the pre-trained AMA model to \textit{casual-adult} and \textit{casual-human} videos, our method performs better. }
  \label{fig8}
\end{figure*}

\noindent\textbf{Texture filtering.} Here we also test the importance of the proposed texture filtering on AMA \cite{vlasic2008articulated}, \textit{casual-adult}, and \textit{casual-human}. The texture rendering results of BANMo \cite{yang2022banmo} and MoDA without texture filtering exhibit noticeable noisy textures on arms (the first row in Figure \ref{fig7}) and human bodies (the second and the third rows in Figure \ref{fig7}). Adding texture filtering to both BANMo and MoDA can effectively alleviate this problem. 
 
\subsection{Motion re-targeting}
We compare BANMo \cite{yang2022banmo} and MoDA’s ability of motion re-targeting. Given the pre-trained model on AMA \cite{vlasic2008articulated} and driving videos of \textit{casual-adult} and \textit{casual-human}, we only optimize the frame-specific camera and body pose codes $\mathbf{\psi}_{c}^{t}$ and $\mathbf{\psi}_{b}^{t}$ while keeping other model parameters unchanged. As shown in Figure \ref{fig8}, BANMo consistently struggles to accurately learn human poses from the driving videos. For instance, BANMo fails to recover the standing pose in the bottom-right of Figure \ref{fig8}, which may stem from its failure to adequately disentangle the shape and pose of the human during optimization. In contrast, MoDA exhibits better performance compared to BANMo.

\section{Conclusion and limitations}
In this paper, we present MoDA, an effective approach for modeling deformable 3D objects from casual videos. We represent 3D objects with a canonical neural radiance field (NeRF) and a deformation model that achieves the 3D point transformation between the observation space and the canonical space. To handle large motions between deformable objects and the background without introducing skin-collapsing artifacts, we propose neural dual quaternion blend skinning (NeuDBS) as our deformation model that can return valid rigid transformations by blending unit dual quaternions. To register 2D pixels across different frames, we model the correspondence learning between canonical feature embeddings of 3D points in the canonical space and 2D image features as an optimal transport problem. Besides, we develop a texture filtering technology for texture rendering that effectively minimizes the impact of noisy colors outside target deformable objects. Extensive experiments on real and synthetic datasets show that the proposed approach can reconstruct 3D shapes for humans and animals with better qualitative and quantitative performance than state-of-the-art methods.

Although MoDA achieves impressive performance in most cases, there are still some limitations that need to be solved in the future. For example, MoDA does not reconstruct the detailed shape of the human hand and the performance on a single video can be further improved. 

\section*{Acknowledgments}

This research is supported by the MoE AcRF Tier 2 grant (MOE-T2EP20220-0007) and the MoE AcRF Tier 1 grant (RG14/22). 

\begin{appendices}

\section{More details of MoDA}\label{secA1}
\begin{figure}[b]
  \centering
  \includegraphics[scale=0.7]{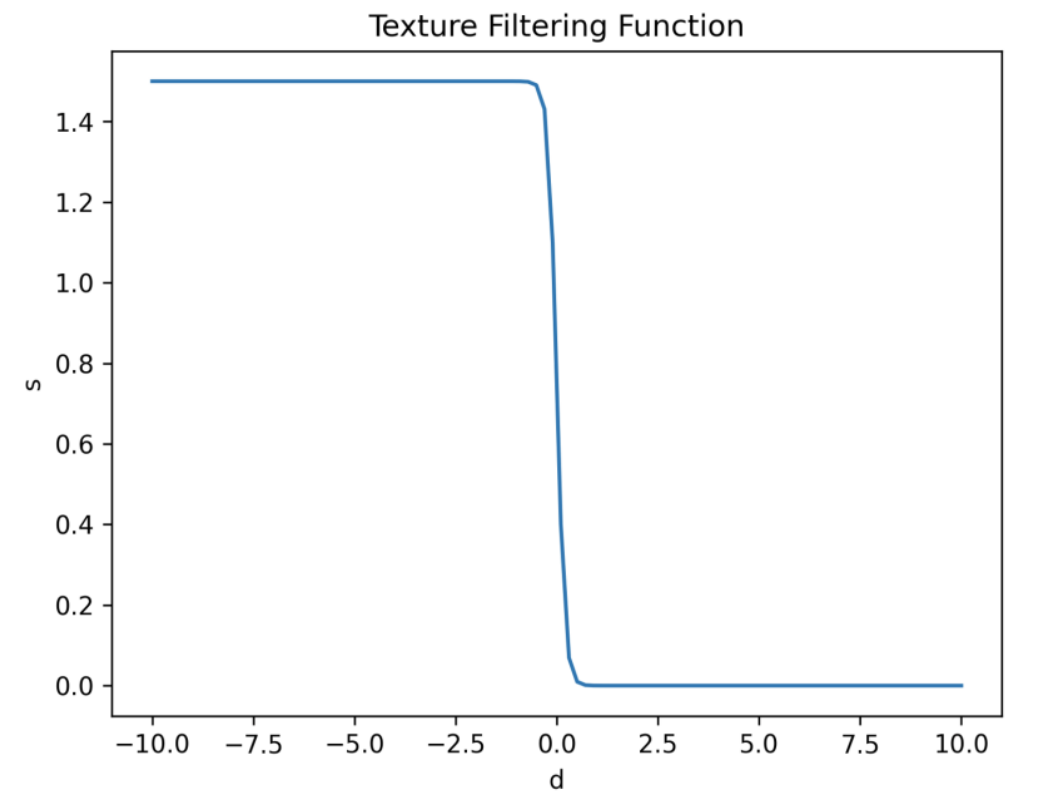}
  \caption{\textbf{The texture filtering function.}}
  \label{supp-func}
\end{figure}
\subsection{Skinning weights for NeuDBS}
We define the skinning weights for NeuDBS as
$\mathbf{W} = \{W_{1},..., W_{J}\} \in \mathbb{R}^{J}$, where $J$ is the number of joint. Learning the skinning weights only from neural networks is difficult to optimize. To obtain the skinning weights for the proposed NeuDBS, we first calculate the Gaussian skinning weights and then learn the residual skinning weights with an MLP network following \cite{yang2022banmo}. 

Firstly, we compute the Gaussian skinning weights based on the Mahalanobis distance between 3D points and the Gaussian ellipsoids, 
\begin{equation}
    \mathbf{W}_{G} = (\mathbf{X}-\mathbf{O})^{T}\mathbf{V}^{T}\boldsymbol{\Lambda}^{0}\mathbf{V}(\mathbf{X}-\mathbf{O}),
\end{equation}
where $\mathbf{O} \in \mathbb{R}^{J \times 3}$ are the joint center locations, $\mathbf{V} \in \mathbb{R}^{J \times 3 \times 3}$ are joint orientations and $\boldsymbol{\Lambda}^{0} \in \mathbb{R}^{J \times 3 \times 3}$ are diagonal scale matrices. The joints represented by explicit 3D Gaussian ellipsoids are composed of these 3 elements: center, orientation, and scale.
To learn better skinning weights for 3D deformation, we predict the residual skinning weights from an MLP network,
\begin{equation}
    \mathbf{W}_{r} = F_{skin}(\mathbf{X}, \boldsymbol{\psi}_{b}),
    \label{res}
\end{equation}
then we have the final skinning weights,
\begin{equation}
    \mathbf{W} = \sigma_{softmax}(\mathbf{W}_{G} + \mathbf{W}_{r}).
\end{equation}
To be specific, the skinning weights $\mathbf{W}^{t}_{o\xrightarrow{}c}$ are learned from 3D points in the observation space and the body pose code $\boldsymbol{\psi}_{b}^{t}$ at time $t$, and $\mathbf{W}^{t}_{c\xrightarrow{}o}$ are learned from 3D points in the canonical space and the rest pose code $\boldsymbol{\psi}_{b}^{*}$.
\subsection{Loss functions}
\noindent\textbf{Optical flow loss.} 
We render 2D flow to compute the optical flow loss. Specifically, we deform the canonical points to another time $t^{\prime}$ and get its 2D re-projection,
\begin{equation}
    \mathbf{x}^{t^{\prime}} = \sum_{k=1}^{N}\tau_{k}\mathbf{P}^{t^{\prime}}(\mathcal{D}^{t^{\prime}}_{c\xrightarrow{}o}(\mathbf{X}^{*}_{k})),
\end{equation}
where $\mathbf{P}^{t^{\prime}}$ is the projection matrix of a pinhole camera. Then we can compute the 2D flow,
\begin{equation}
    \mathbf{f}(\mathbf{x}^{t}, t\xrightarrow{}t^{\prime}) = \mathbf{x}^{t^{\prime}} - \mathbf{x}^{t},
\end{equation}
and the optical flow loss $\mathcal{L}_{of}$ is defined as
\begin{equation}
\label{of}
    \mathcal{L}_{of} = \sum_{\mathbf{x}^{t}, (t, t^{\prime})}{\left\|\mathbf{f}(\mathbf{x}^{t}, t\xrightarrow{}t^{\prime}) - \widetilde{\mathbf{f}}(\mathbf{x}^{t}, t\xrightarrow{}t^{\prime})\right\|^{2}}, 
\end{equation}
where $\widetilde{\mathbf{f}}$ is the observed optical flow that are extracted from off-the-shelf method \cite{yang2019volumetric}.

\noindent\textbf{3D cycle consistency loss.}
Similar to \cite{li2021neural, yang2022banmo}, we introduce a 3D cycle consistency loss to learn better deformations. We deform the sampled points in the observation space to the canonical space and then deform them back to their original coordinates, 
\begin{equation}
    \mathcal{L}_{cyc} = \sum_{k}\tau_{k}{\left\| \mathcal{D}^{t}_{c\xrightarrow{}o}(\mathcal{D}^{t}_{o\xrightarrow{}c}(\mathbf{X}^{t}_{k})) - \mathbf{X}^{t}_{k})\right\|^{2}_{2}}, 
\end{equation}
where $\tau_{k}$ weighs the sampled points to guarantee the points closer to the surface have stronger regularization.

\noindent\textbf{Eikonal loss.} 
Following \cite{yang2022banmo, yang2023reconstructing}, we also adopt the implicit geometric regularization term \cite{gropp2020implicit} as :
\begin{equation}
    \mathcal{L}_{eikonal} = \sum_{\mathbf{X} \in \mathbf{V}^{*}}(\left\|\nabla F_{\mathrm{SDF}}(\mathbf{X})\right\|_{2} - 1)^{2}.
\end{equation}

\subsection{Implementation details}
\noindent\textbf{Training strategy.} The optimization strategies of MoDA include three stages. Firstly, we optimize all losses and parameters. In this stage, MoDA already reconstructs good shape and deformation. Then we improve the articulated motions, where we only update the parameters related to the deformation model while keeping the shape parameters fixed. Finally, we improve the details of the reconstructions through importance sampling while freezing the camera poses. The design of MLP networks in MoDA is similar to BANMo \cite{yang2022banmo}. The hyperparameters $\gamma$ and $\lambda$ in the texture filtering function are set to $1.5$ and $10$ respectively (See Figure \ref{supp-func} for the function). Our code will be available on GitHub once the paper is accepted.

\noindent\textbf{Sampling details for 2D-3D matching.} In optimal transport, the sampling of 3D points is similar to BANMo \cite{yang2022banmo}. We establish a canonical 3D grid $\mathbf{V}^{*} \in \mathbb{R}^{20\times20\times20}$ ($N_{point}= 8000$) to build correspondence between pixels and canonical points. This grid is centered at the origin and axis-aligned with bounds $[x_{min}, x_{max}], [y_{min}, y_{max}]$, and $[z_{min}, z_{max}]$, undergoes iterative refinement during optimization. Every 200 iterations, we update the bounds of the grid by approximating the object's bounds. This approximation is obtained by applying marching cubes on a $64^{3}$ grid to extract a surface mesh.

\subsection{Dataset}\label{dataset}
We use 6 datasets in this work, where \textit{casual-cat}, \textit{casual-human}, \textit{eagle} and \textit{hands} are collected by BANMo \cite{yang2022banmo}. We have obtained permission to use these datasets. AMA dataset is the published dataset collected by \cite{vlasic2008articulated}. The usage of \textit{casual-adult} has also obtained consent.

\section{More experimental results}
\begin{figure*}[h]
  \centering
  \includegraphics[scale=0.35]{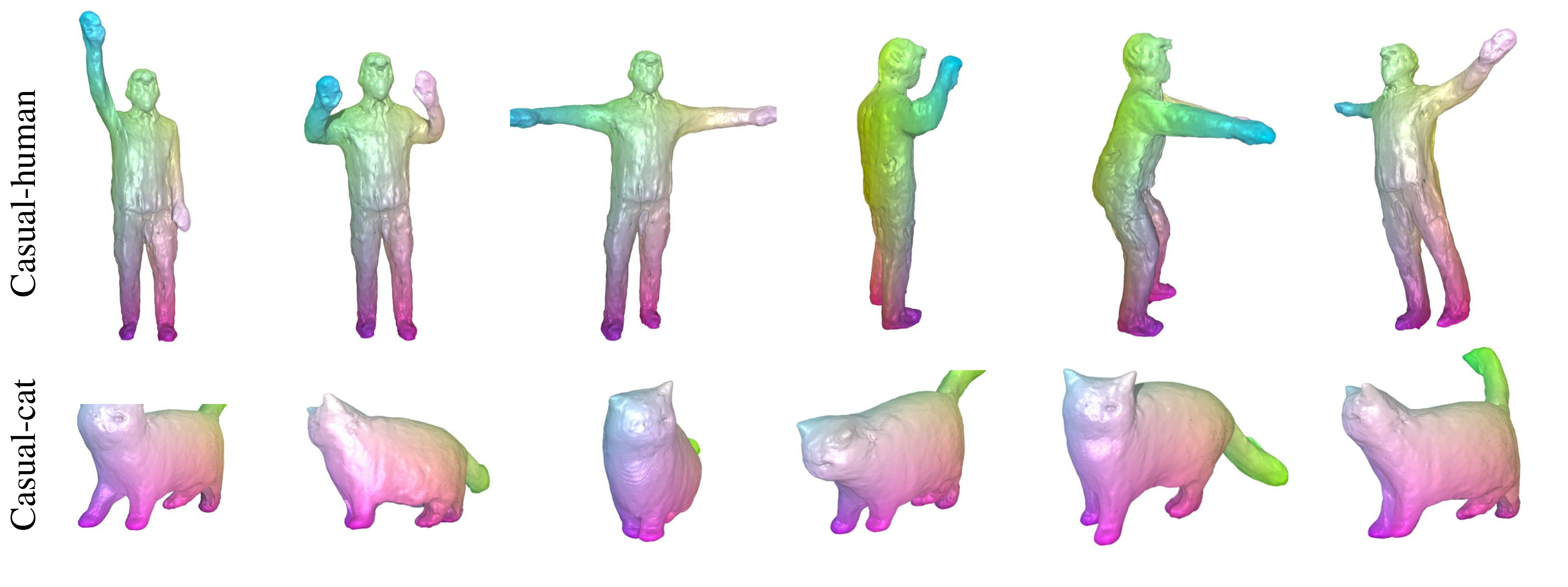}
  \caption{\textbf{Correspondence between different videos on \textit{casual-human} and \textit{casual-cat}.} Distinct colors represent the correspondence.}
  \label{supp-fig2}
\end{figure*}

\begin{figure*}[h]
  \centering
  \setlength{\abovecaptionskip}{0.cm}
  \includegraphics[scale=0.4]{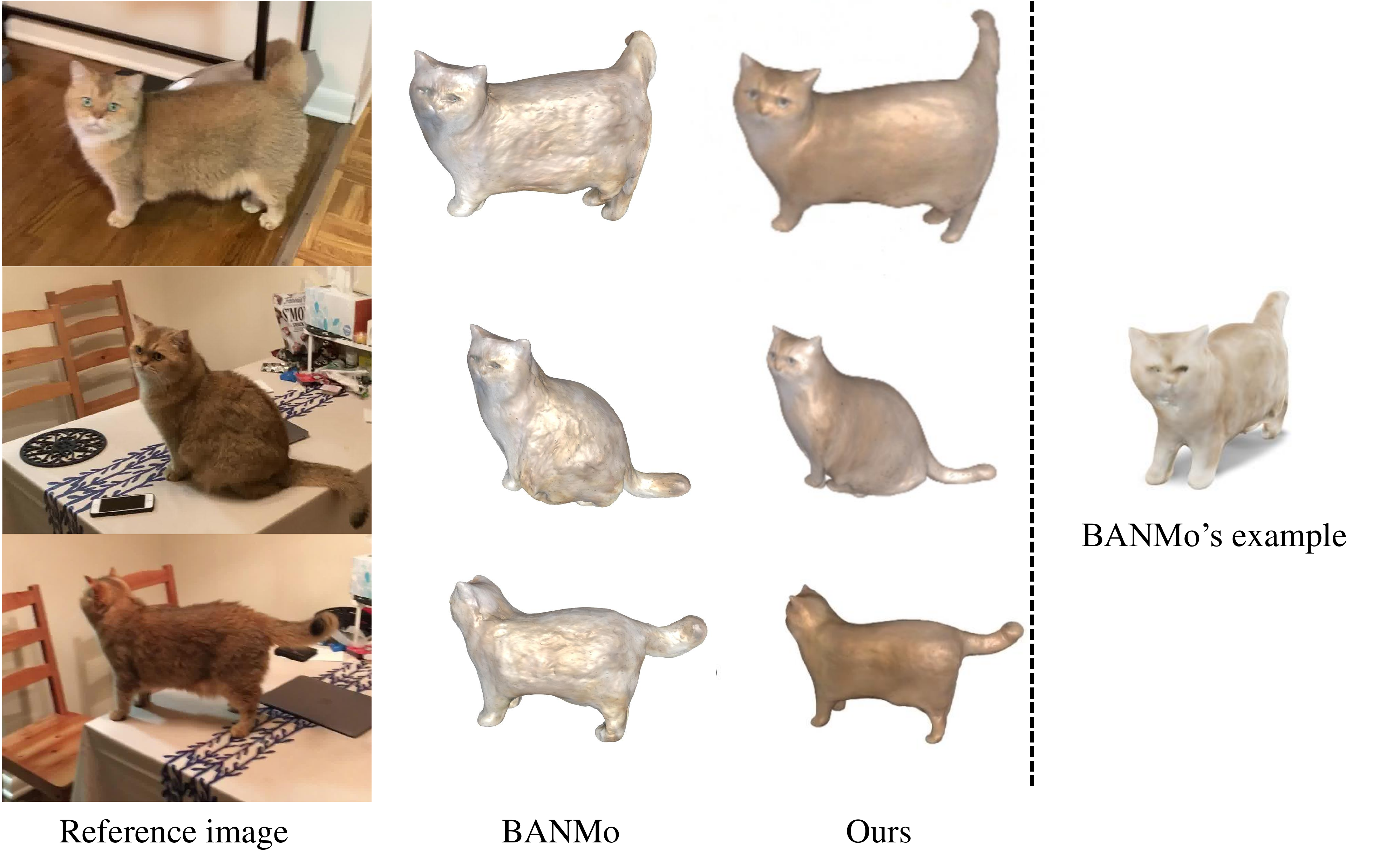}
    \caption{\textbf{Texture comparison of BANMo and our method on casual-cat.} The fourth column is a screenshot of the example provided on BANMo's website.}
      \label{supp-fig4}
\end{figure*}
In this section, we show more experimental results.

\noindent\textbf{Correspondence.} In Figure \ref{supp-fig2}, we illustrate the correspondence between different videos in both the \textit{casual-human} and \textit{casual-cat} datasets. Distinct colors represent the correspondence. 

\noindent\textbf{Texture rendering results.} Here, we compare texture rendering results of BANMo \cite{yang2022banmo} and our method on \textit{casual-cat}. As shown in \ref{supp-fig4}, we also provide a screenshot of the example on BANMo's website\footnote{https://banmo-www.github.io}. The result of BANMo is aligned with the example on their website, appears unrealistic and potentially influenced by noise. In contrast, our method produces results more closely resembling the reference image. 

\begin{figure*}
  \centering
  \includegraphics[scale=0.14]{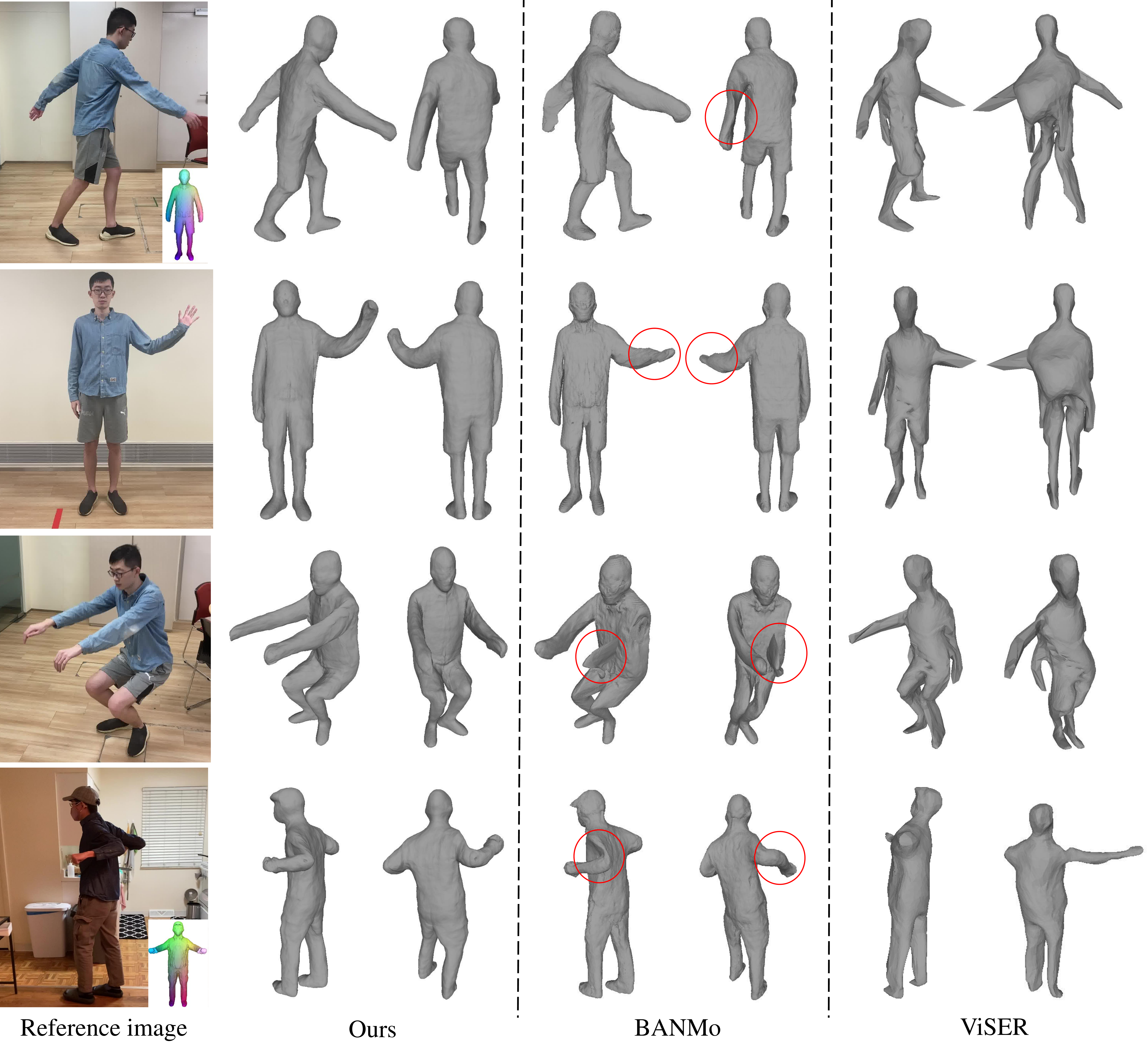}
  \caption{\textbf{Qualitative comparison of different methods on \textit{casual-human} and \textit{casual-adult} (multiple-video setups).} We show 2 views of the reconstruction results based on the reference images. ViSER \cite{yang2021viser} fails to learn detailed 3D shapes and accurate poses from the videos. BANMo \cite{yang2022banmo} has obvious skin-collapsing artifacts (in the red circles) for motions with large joint rotations while our method performs well. The rest poses of \textit{casual-human} and \textit{casual-adult} are shown in the lower right corner.}
  \label{supp-fig3}
\end{figure*}

\noindent\textbf{More results on multiple-video setups.} We also show more results comparing MoDA with BANMo \cite{yang2022banmo} and ViSER \cite{yang2021viser} on \textit{casual-human} and \textit{casual-adult} in Figure \ref{supp-fig3}. 

\section{Training time}
\begin{table}[t]
  \centering
  \caption{\textbf{Training time comparison.} We compare the training times of our method and BANMo \cite{yang2022banmo} on different datasets. We also show the number of videos and frames of different datasets for reference. The unit of time is hour.}
  \label{time}
  \begin{tabular}{ccccc}
    \toprule
  \multirow{2}{*}{Dataset}  & \multirow{2}{*}{Video} & \multirow{2}{*}{Frame} & \multicolumn{2}{c}{Time} \\ \cmidrule{4-5}
  & &  & BANMo &MoDA  \\
    \midrule
  AMA & 16 & 2600 & 10.00 &11.00 \\
  casual-cat& 11  & 900 & 8.75 & 9.50\\
   casual-human &  10& 584 & 8.00 & 9.00  \\
   casual-adult & 10 & 1000 &   9.00 & 10.25\\
    eagle &  5& 750 & 8.25 & 9.25 \\
    hands&  5& 750 & 8.25 & 9.25\\
    \bottomrule
  \end{tabular}
\end{table}
We compare the training times of our method and BANMo \cite{yang2022banmo} on different datasets. We train the models on two RTX 3090 GPUs. As shown in Table \ref{time}, MoDA and BANMo both have fast training on different datasets. BANMo takes around 8-10 hours. MoDA takes about one hour more than BANMo. MoDA requires more computational time compared to BANMo due to two primary factors. Firstly, DBS employed in MoDA is inherently more time-consuming than Linear Blend Skinning (LBS), as reported in Figure 18 of \cite{kavan2008geometric}. Secondly, the optimization process for solving the optimal transport problem adds additional computational overhead. However, the increased training time is acceptable considering MoDA's superior performance.




\end{appendices}

{\small
\bibliographystyle{ieee_fullname}
\bibliography{egbib}
}

\end{document}